\documentclass[conference]{IEEEtran}
\pdfoutput=1
\usepackage{times}
\usepackage{epsfig}
\usepackage{graphicx}
\usepackage{amsmath}
\usepackage{amssymb}
\usepackage{multirow}
\usepackage{tablefootnote}
\usepackage{booktabs}
\usepackage{threeparttable}
\usepackage{subfig}
\usepackage[export]{adjustbox}
\graphicspath{{./figures/}}
\usepackage{amssymb}
\usepackage{pifont}
\usepackage{footmisc}
\newcommand{\cmark}{\ding{51}}%

\usepackage{tikz}
\usepackage{url}
\usepackage{pdfpages}
\usepackage{afterpage}
\usepackage{color, colortbl}
\usepackage{import}
\usepackage{algorithm,algpseudocode}
\usepackage[titles,subfigure]{tocloft}
\usepackage{changepage}
\usepackage{makecell}
\usepackage[misc]{ifsym}

\makeatletter
\DeclareRobustCommand*\textsubscript[1]{%
  \@textsubscript{\selectfont#1}}
\def\@textsubscript#1{%
  {\m@th\ensuremath{_{\mbox{\fontsize\sf@size\z@#1}}}}}
\makeatother
\IEEEoverridecommandlockouts

\usepackage[pagebackref=true,breaklinks=true,colorlinks,bookmarks=false]{hyperref}

\begin{document}

\title{EDNet: Efficient Disparity Estimation with Cost Volume Combination and Attention-based Spatial Residual}

\author{\IEEEauthorblockN{Songyan Zhang\IEEEauthorrefmark{2}, Zhicheng Wang\IEEEauthorrefmark{1}\thanks{*Corresponding author.}\IEEEauthorrefmark{2}, Qiang Wang\IEEEauthorrefmark{3}, Jinshuo Zhang\IEEEauthorrefmark{2}, Gang Wei\IEEEauthorrefmark{2}, Xiaowen Chu\IEEEauthorrefmark{3}\\}
\IEEEauthorblockA{\IEEEauthorrefmark{2}CAD Research Center, Tongji University. \\
\IEEEauthorrefmark{3}Department of Computer Science, Hong Kong Baptist University.\\
\{spyder, zhichengwang, zhangjinshuo, weigang\}@tongji.edu.cn}
\{qiangwang, chxw\}@comp.hkbu.edu.hk}

\maketitle

\begin{abstract}
   Existing state-of-the-art disparity estimation works mostly leverage the 4D concatenation volume and construct a very deep 3D convolution neural network (CNN) for disparity regression, which is inefficient due to the high memory consumption and slow inference speed. In this paper, we propose a network named EDNet for efficient disparity estimation. Firstly, we construct a combined volume which incorporates contextual information from the squeezed concatenation volume and feature similarity measurement from the correlation volume. The combined volume can be next aggregated by 2D convolutions which are faster and require less memory than 3D convolutions. Secondly, we propose an attention-based spatial residual module to generate attention-aware residual features. The attention mechanism is applied to provide intuitive spatial evidence about inaccurate regions with the help of error maps at multiple scales and thus improve the residual learning efficiency. Extensive experiments on the Scene Flow and KITTI datasets show that EDNet outperforms the previous 3D CNN based works and achieves state-of-the-art performance with significantly faster speed and less memory consumption.
\end{abstract}


\section{Introduction}

Accurate and fast depth estimation is of great significance to many applications like robot navigation, 3D reconstruction and autonomous driving. Instead of depth regression from a single-view RGB image, stereo matching is to conduct correspondence analysis between pixels of stereo images and compute the disparity $d$ for each pixel. Depth can be then calculated by $(\frac{fB}{d})$, where $f$ is the camera’s focal length and $B$ is the distance between two camera centers, also called baseline in stereo vision.

While traditional methods based on hand-crafted feature extraction and matching cost aggregation tend to fail on those textureless and repetitive regions in the images, convolutional neural networks (CNNs) have been widely adopted to conquer those difficulties in stereo matching. Several recent methods \cite{gcnet,psmnet,gwcnet,ganet} have achieved state-of-the-art performance by constructing a 4D concatenation volume which follows 3D convolution blocks for aggregation. Although the 4D concatenation cost volume can preserve the rich contextual information in conjunction with the strong regularization ability of 3D convolutions, it significantly increases the computation cost and usually cannot perform real-time disparity inference. Moreover, the concatenation volume incorporates no feature similarity measurement, which means that the model has to learn correspondence from scratch. Besides, DispNetC \cite{dispnetc} formed a low-cost correlation layer with 2D convolutions to conduct correspondence analysis between the left and right feature maps. The following works \cite{consistency,segstereo,CRL} adopted the similar method as it keeps a good balance between speed and accuracy. However, as the correlation map is produced with only one single feature channel for each disparity level, the performance is less competitive. Thus, this raises the question of how to make full use of the complementary advantages of the concatenation volume and correlation volume.
\begin{figure*}[htbp]
	\centering
	\includegraphics[width=0.94\linewidth]{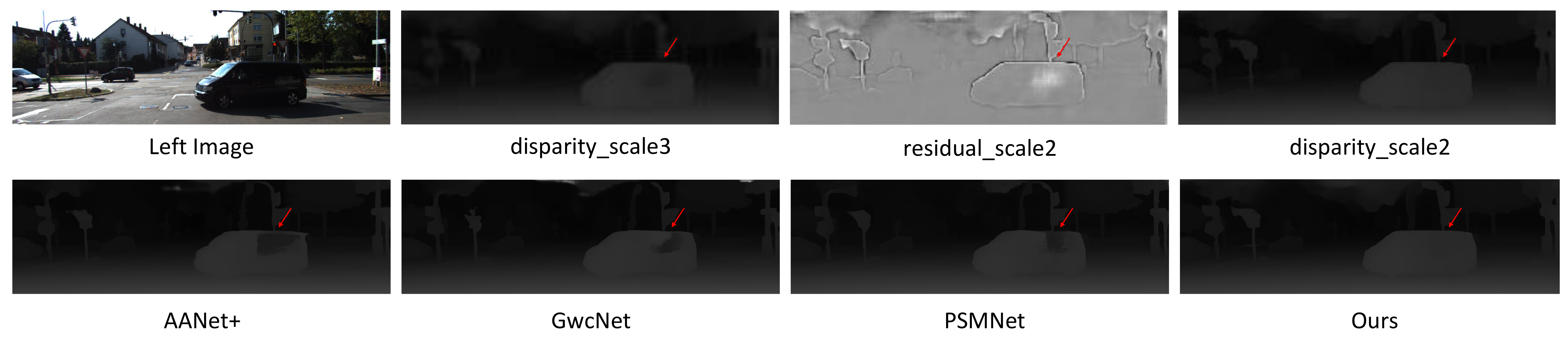}
	\caption{The first row is the visualization of residual learning process from scale 3 to scale 2. The residual\_scale2 is learned from the disparity\_scale3  for correcting the disparity\_scale2. With our proposed modules, sharp edges and overall structures can be recovered. Other state-of-the-art methods fail to generate the accurate disparity estimation in low-texture regions as shown in the second row. Please pay more attention to regions pointed by the red arrow.}
	\label{fig:residual}
	\vspace{-0.8 em}
\end{figure*}

Since ResNet \cite{resnet} has revealed that the residual convolution block can improve the training efficiency by learning a residual mapping instead of the desired underlying one, it has been adopted as a popular approach to refine the disparity estimation \cite{wang2020fadnet,edgestereo,consistency,AANET}. To be specific in stereo matching, learning an additive correction to the coarse disparity map is easier and more efficient than directly learning the fine-grained one. However, some works failed to provide the residual learning module with the fitting error information \cite{madnet,geonet}, or computed the estimated error at only the one scale \cite{wang2020fadnet,CRL,flownet2} but adopted it to learn the disparity maps at multiple scales. 
The error map from one single scale cannot provide the precise error information at other scales, which makes the residual learning method less effective.
Furthermore, even if the error map is provided at each corresponding scale \cite{edgestereo}, the conventional residual learning method has no explicit spatial guidance about where to intervene. As the regions with inaccurate estimation deserve more attention, we argue that residual learning could be more efficient if the spatial attention about the learning errors is provided.

To address the above issues, we propose EDNet which is composed of a combined volume to generate robust feature representations, and an attention-based residual module to learn the disparity refinement. Firstly, the proposed combined volume alleviates the information loss by employing the squeezed concatenation volume and preserves the feature similarity measurement with the correlation volume. We then adopt 2D convolutions for further aggregation so that the significant memory consumption and computational complexity of 3D CNNs can be avoided. Secondly, inspired by the attention mechanism, we adopt a spatial attention module to generate the attention-aware residual features. Therefore, the residual learning module can have intuitive spatial evidence about inaccurate regions to compute a specific correction. We follow the coarse-to-fine strategy and compute the attention-aware residuals across different scales. With the error maps provided at each scale, the residual module can learn a corresponding correction accordingly and improve the learning efficiency. As shown in Figure \ref{fig:residual}, our network can generate an accurate and continuous disparity map even in low-texture regions. The contributions of our work can be summarized as follows:

\begin{itemize}
    \item We propose a low-cost but effective method to aggregate the 3D correlation features and 4D concatenation volume together by constructing a combined volume, which can be further processed by fast 2D convolutions. Compared with others, our correspondence analysis can preserve both the contextual information and feature similarities even with 2D convolutions.
    \item We design an Attention-based Residual (AR) module to learn the disparity refinement at each scale. In the AR module, the attention mechanism is applied to the concatenated maps of RGB image, estimated disparity and estimated error to improve the learning efficiency.
    \item Compared to those existing methods based on 3D CNNs, our proposed EDNet achieves state-of-the-art accuracy on the public Scene Flow \cite{dispnetc} and KITTI \cite{KITTI2015,KITTI2012} datasets with up to 76.5\% runtime memory reduction and 45$\times$ inference speed acceleration.
\end{itemize}

The rest of the paper is organized as follows. We introduce some related studies about stereo matching based on CNNs in Section \ref{sec:related_work}. Section \ref{sec:model} introduces the methodology and implementation of our proposed EDNet. We demonstrate our experimental results in Section \ref{sec:exp}. We finally conclude the paper in Section \ref{sec:conclusion}.
\section{Related Works}\label{sec:related_work}

A classical stereo matching pipeline consists of four steps \cite{taxonomy2001}. In recent years, CNNs have drawn great attention and been introduced to tackle the stereo matching task \cite{LeCun2016}. In this section, we briefly discuss those common mechanisms for computing the matching cost with CNNs and review the approaches with the residual learning method.

\subsection{Matching Cost Computation}
CNN based matching cost computation methods make a great contribution to the stereo matching accuracy. There are two popular approaches for matching cost computation. The first one is using either a layer of 2D \cite{flownet} or 1D \cite{dispnetc} convolutional operations, called correlation layer. Such an inner product between feature vectors is adopted in \cite{embedding2015,feng2017,consistency,segstereo}. Liang $et$ $al.$ \cite{consistency} build a correlation volume for initial disparity estimation which follows a disparity refinement module by learning through feature consistency. Wang $et$ $al$. \cite{wang2020fadnet} make some modification and propose a point-wise correlation volume to preserve fast computation. Another popular method to compute matching cost is to form a 4D volume by concatenating the corresponding features from the opposite stereo images across each disparity level. 3D convolutions are followed to aggregate features and regress disparity. This method can be found first in \cite{gcnet}. Chang $et$ $al$. \cite{psmnet} improve Kendall’s approach \cite{gcnet} by designing a spatial pyramid pooling module \cite{SPP} so that correspondence estimation can benefit from the image features with rich object context information. Guo $et$ $al$. \cite{gwcnet} combine the concatenation volume with the group-wise correlation volume and improve the accuracy with 3D convolutions. The best performance on Scene Flow dataset comes from \cite{MCUS} which introduces the idea of DenseNet \cite{densenet} to further improve PSMNet \cite{psmnet}. Zhang $et$ $al$. propose GANet \cite{ganet} 
with two guided aggregation layers and fifteen 3D convolutions to achieve state-of-the-art performance.

\subsection{Residual Learning for Stereo Matching}
The residual learning concept is proposed by He $et$ $al$. \cite{resnet} which turns to be an efficient way to train a CNN model and has been adopted by many works. In the stereo matching task, the residual learning strategy is widely used for refining disparity estimation \cite{consistency,wang2020fadnet,stereonet,AANET}. Pang $et$ $al$. \cite{CRL} present a cascade residual learning scheme and adopt a two-stage CNN, in which the second stage refines the estimation by producing residual signals. Stucker $et$ $al$. \cite{resdepth} specially build a U-Net \cite{unet} based network to enhance the reconstruction by regressing a residual correction. In order to meet the need for real-time inference, \cite{anytime} takes residual learning strategy to flexibly output disparity estimation according to the requirement of applications. Song $et$ $al$. \cite{edgestereo} manage to aggregate edge information for residual learning and thus construct a multi-task network for edge detection and stereo matching.
\begin{figure*}[htbp]
	\centering
	\includegraphics[width=0.92\linewidth]{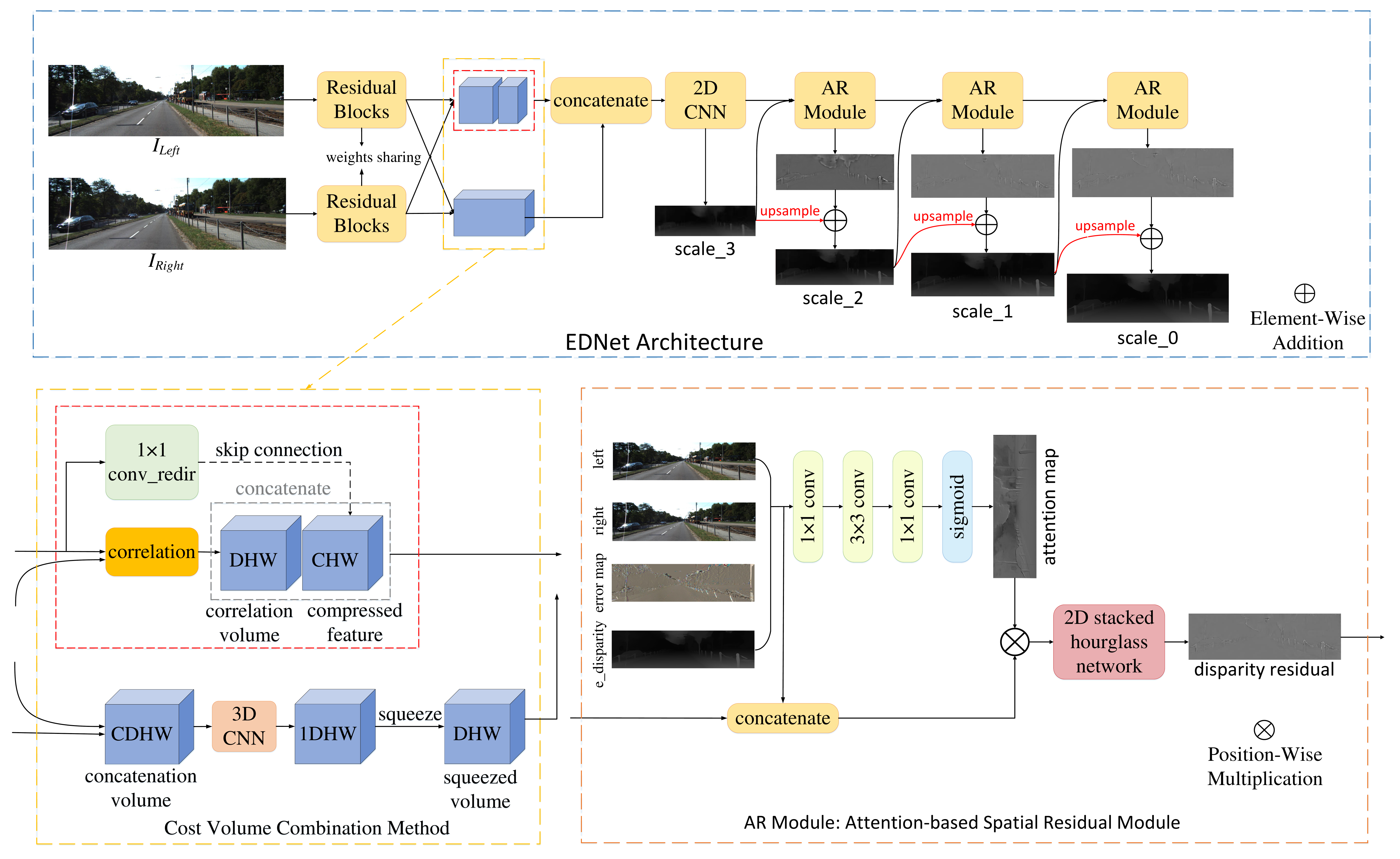}
	\caption{An overview of our proposed EDNet. We exploit the architecture of DispNetC as the backbone. The attention-based spatial residual module and combined volume are proposed for accuracy and efficiency improvement. In order to provide better visualization, the skip connections between the encoder and decoder networks in DispNetC and some other data flow are omitted here.}
	\label{fig:ednet}
	\vspace{-0.8 em}
\end{figure*}
\section{Methodology}\label{sec:model}

\subsection{Network Architecture}
The architecture of our proposed EDNet is shown in Figure \ref{fig:ednet}. We exploit the structure of DispNetC \cite{dispnetc} as the backbone with extensive modifications. For feature extraction, the last left and right feature maps of conv3 from the weights-sharing encoder network are used to form the combined volume which is composed of a squeezed concatenation volume and correlation volume which will be discussed in Section \ref{subsec:cost_volume}. The detail of our proposed cost volume combination method can be found in the left bottom corner of Figure \ref{fig:ednet}. 2D convolutions are then employed to aggregate the combined volume and regress the disparity. In the decoder part, we follow the coarse-to-fine strategy to refine the disparity progressively. The spatial attention module is applied in order to generate attention-aware residual features, which will be introduced in Section \ref{subsec:attention}. The stacked hourglass module in PSMNet \cite{psmnet} is used for residual regression but is implemented by 2D convolutions. The attention-based spatial residual module is well illustrated in the right bottom corner of Figure \ref{fig:ednet}. Different from DispNetC \cite{dispnetc} which has 6 scales of output, we reduce the disparity prediction to 4 scales, removing the prediction at 1/16 and 1/32 of full resolution.

\subsection{Cost Volume Combination}\label{subsec:cost_volume}
Previous works simply build a correlation volume \cite{dispnetc,CRL,wang2020fadnet} or 4D concatenation volume \cite{gcnet,gwcnet,psmnet} which follows 2D or 3D convolutions for aggregation. However, a single cost volume cannot meet the need of preserving contextual information and feature similarity at the same time. GwcNet \cite{gwcnet} improves the performance by combining the group-wise correlation and concatenation volume, but 3D convolutions are required for aggregation which leads to higher memory consumption and more complex computation. To this end, we propose to combine the correlation volume and 4D concatenation volume in a more efficient way aiming at taking advantages of both two cost volumes.

Given a pair of stereo features $\textbf{f}_L$ and $\textbf{f}_R$, we follow the 1D-correlation in DispNetC \cite{dispnetc} to calculate the correspondence at each disparity level $d$. The correlation volume is computed as:
\begin{align}
    \textbf{C}_{corr}(d, x, y) = \frac{1}{N}<\textbf{f}_L(x-d, y), \textbf{f}_R(x, y)>
\end{align}

where $<x_1,x_2>$ is the inner product of two feature vectors $x_1$ and $x_2$, and $N$ is the channel number of input features. 
The shape of the correlation volume is $N \times D \times H \times W$, where $N$ denotes the batch size, $D$ is the estimated disparity range and the spatial size is $H \times W$. Then we construct the 4D concatenation volume by concatenating the left and right feature maps, i.e.,
\begin{align}
    \textbf{C}_{concat}(d, x, y) = Concat\{\textbf{f}_L(x-d, y), \textbf{f}_R(x, y)\}
\end{align}

After obtaining the concatenation volume with the shape $N \times D \times C \times H \times W$, we use three 3D convolutions for aggregation and compress it into 1 channel. The aggregated concatenation volume now has the shape $N \times D \times 1 \times H \times W$. It is then squeezed into $N \times D \times H \times W$, the same shape as the correlation volume. The correlation volume and squeezed concatenation volume are finally concatenated to form the combined volume. In this way, both the contextual information and feature similarity measurement are incorporated in the combined volume. Further aggregation can be done by 2D convolutions instead of 3D convolutions, which are more efficient. The final combined volume is formed as:
\begin{align}
    \textbf{C}_{comb}(x, y) = Concat\{\textbf{C}_{corr}(x, y), \textbf{C}_{concat}(x, y)\}
\end{align}

\subsection{Attention-based Spatial Residual}\label{subsec:attention}
The normal residual learning method lacks the spatial evidence about where the errors occur. We propose an attention-based  spatial residual module to guide the residual learning process to pay more attention to those inaccurate regions across the whole spatial space. According to the estimated disparity $\hat{d}^s$ at scale $s$, a synthesized left image $\tilde{I}_{L}^s$ can be obtained by warping the right image $I_{R}^s$, i.e.,
\begin{align}
    \tilde{I}_{L}^s(x, y)=I_{R}^{s}(x+\hat{d}^s(x, y), y)
\end{align}

With the warped left image and target left image, we can get the error $E_{L}^s=|\tilde{I}_{L}^s-I_{L}^s|$. A spatial attention module with 3 layers of 2D convolution which are 1$\times$1, 3$\times$3 and 1$\times$1 respectively is applied. The spatial attention feature map $\textbf{f}_{a}^s$ is compressed into one channel followed by the sigmoid function to compute the spatial attention vector whose size is $N \times 1 \times H \times W$. The input of error map and color stereo images enables the spatial attention module to learn an attention distribution on blurry object boundaries and mismatched pixels. Akin to FADNet \cite{wang2020fadnet}, CRL\cite{wang2020fadnet}, FlowNet2 \cite{flownet2}, the input to spatial attention module is the concatenation of the stereo images, error map and estimated disparity map.

We follow DispNetC \cite{dispnetc} to preserve both the high-level information and local information by skip-connection. The concatenation of `upconvolution' feature maps from the decoder network and corresponding feature maps from the encoder network is then concatenated with the input of attention module to form the residual feature maps $\textbf{f}_{r}^s \in R^{H \times W \times C}$. The final attention-aware residual features $\textbf{f}_{ar}^s \in R^{H \times W \times C}$ at scale $s$ are computed by multiplying the attention vector and residual features  $\textbf{f}_{r}^s$, i.e.,
\begin{align}
    \textbf{f}_{ar}^s=\textbf{f}_{ar}^s \otimes \sigma(\textbf{f}_{a}^s)
\end{align}
where $\sigma(\cdot)$ denotes the sigmoid function. The attention-aware residual features are then input to the stacked hourglass module for residual regression. The stacked hourglass module has the same encoder-decoder structure as PSMNet \cite{psmnet} but is implemented with 2D convolutions.

The attention-based spatial residual module is repeated 3 times as we increase the resolution of the disparity map progressively. Different from the aforementioned works \cite{wang2020fadnet, CRL}, we compute error maps across multiple scales as the error information changes accordingly after the correction. Therefore, we provide the residual learning module with constantly updated error maps across multiple scales instead of a single error map at full resolution.

\subsection{Multi-scale Residual Learning}
Instead of building a cascade architecture with residual refinement at the second stage \cite{flownet2, CRL,wang2020fadnet}, we simply replace the direct disparity estimation with a residual estimation over all scales except the smallest $S$ at which the initial disparity is computed. The multi-scale residual outputs are denoted as $\{r^s\}_{s=0}^{S-1}$ where 0 represents the scale of full resolution. For the rest $S$ scales, the estimated disparity at the previous scale is first upsampled to the current scale $\hat{d}_{up}^s$ using bilinear interpolation function and then added to the residual for refinement. The final predicted disparity $\hat{d}^s$ at scale $s$ is produced as:
\begin{align}
    \hat{d}^s=\hat{d}_{up}^s+r^s, 0 \leq s \leq S-1
\end{align}

\subsection{Loss Function}
Given the output disparity map at different scales, we adopt the pixel-wise smooth L1 loss to train our EDNet at scale $s$:
\begin{align}
    L^s(d^s, \hat{d}^s)=\frac{1}{N}\sum_{i=1}^{N}{smooth}_{L_1}(d_{i}^s - \hat{d}_{i}^s), \label{eq:smooth_l1}
\end{align}
where $N$ is the number of pixels of the disparity map, $\hat{d}_{i}^s$ is the $i^{th}$ element of the predicted disparity $\hat{d}^s$, and $d^s$ represents the ground truth disparity. The smooth L1 loss function is:
\begin{align}
    {smooth}_{L_1}(x)=
    \begin{cases}
    0.5x^2,& \text{if } |x| < 1\\
    |x|-0.5,              & \text{otherwise}.
\end{cases}
\end{align}

The final loss function is a combination of losses over all scales, i.e.,
\begin{align}
    L = \sum_{s=0}^{S} {\lambda}^s L^s (d^s, \hat{d}^s)
\end{align}
where ${\lambda}^s$ is a scalar for adjusting the loss weight at scale $s$.
\section{Performance Evaluation}\label{sec:exp}
\subsection{Datasets and Evaluation Metrics}
Three public datasets are used for training and testing our EDNet. The Scene Flow dataset \cite{dispnetc} consists of 39,824 pairs of synthetic stereo RGB images (35,454 pairs for training and 4,370 pairs for testing) with a full resolution of 960$\times$540. Both KITTI 2012 \cite{KITTI2012} and KITTI 2015 \cite{KITTI2015} are datasets of real scenes with a full resolution of 1242$\times$375. The ground truth of these two datasets is generated by lidar so that only sparse ground truth is available. We evaluate our model on Scene Flow dataset with the end-point error (EPE), 1-pixel error and 3-pixel error. The end-point error computes the mean disparity error in pixels while the 1-pixel error and 3-pixel error measure the average percentage of the pixel whose EPE is bigger than 1 pixel and 3 pixels respectively. The official metrics (e.g., D1-all) are reported for evaluation on KITTI 2012 and KITTI 2015 datasets.
\begin{table*}[!ht]
		\centering
		\begin{tabular}{cccccccc} \hline
			\multirow{2}{*}{Method} & \multicolumn{2}{c}{Cost Volume} & \multicolumn{2}{c}{Residual Module} & \multicolumn{3}{c}{Results} \\ 
			& correlation & s-concatenation & normal & attention & EPE & $\textgreater$1px(\%) & $\textgreater$3px(\%) \\ \hline
			EDNet-NRS & \cmark & & & & 1.67 & 27.9 & 9.7 \\ \hline
			EDNet-NRCo  &  & \cmark & & & 1.89 & 29.9 & 10.7 \\ \hline
			EDNet-NR  & \cmark & \cmark & & & 1.63 & 27.4 & 9.8 \\ \hline
			EDNet-NA  & \cmark & \cmark & \cmark & & 1.04 & 12.7 & \textbf{5.4} \\ \hline
			EDNet-NS  & \cmark & & & \cmark & 1.07 & 13.1 & 5.8 \\ \hline
			EDNet-F   & \cmark & \cmark & & \cmark & \textbf{1.00} & \textbf{12.2} & \textbf{5.4} \\ \hline
		\end{tabular}
        \caption{Evaluation of EDNet with different settings. We compute the end-point error, 1-pixel and 3-pixel error on Scene Flow dataset. We use "Co", "S", "R", "A" to denote the correlation volume, squeezed concatenation volume, normal residual, and attention-based residual respectively. "N" stands for "not applied". "EDNet-F" represents the model with all our proposed components and "s-concatenation" means the squeezed concatenation volume.}
		\label{tab:ablation}
\end{table*}

\subsection{Implementation Details}
We implemented our EDNet by PyTorch\cite{pytorch} and trained the model with Adam (momentum=0.9, beta=0.999) as optimizer. For Scene Flow dataset, raw images are randomly cropped to 320$\times$640 as input. Our training is performed on 2 NVIDIA RTX 2080ti GPUs for 70 epochs with a batch size of 8 (4 on each GPU). We follow the training strategy in AANet \cite{AANET}, where the initial learning rate is set to 0.001 and decreased by half every 10 epochs after 20$^{th}$ epoch. The loss weights are set to ${\lambda}_0=1.0, {\lambda}_1={\lambda}_2=0.8, {\lambda}_3=0.6$. The crop size for KITTI is set as 256$\times$512. Due to the insufficient training samples of both KITTI 2012 and KITTI 2015, the pre-trained Scene Flow model is used for fine-tuning on the mixed KITTI 2012 and KITTI 2015 training sets for the first 1000 epochs which follows another 400 epochs of training on two datasets to get the submission result respectively. We use a constant learning rate of 0.0001 for KITTI datasets. Inspired by \cite{nature2003} that searching the correspondence at a coarse scale can be beneficial, especially in low-texture or textureless regions, the loss weights are set as ${\lambda}_0=0.6, {\lambda}_1={\lambda}_2=0.8, {\lambda}_3=1.0$. For all datasets, color normalization is taken into use with the mean ([0.485, 0.456, 0.406]) and variation ([0.229, 0.224, 0.225]) of the ImageNet \cite{ImageNet} for data pre-processing. The maximum disparity is set as 192 pixels.

\subsection{Ablation Study}
To validate the effectiveness of each component in EDNet, we evaluate our model with different configurations on Scene Flow dataset. All the experimental results are obtained after 10 epochs of training.  As shown in Table \ref{tab:ablation}, removing either of the combined volume or attention-based spatial residual leads to a clear drop of performance.
\begin{figure}[htbp]
	\centering
	\includegraphics[width=0.9\linewidth]{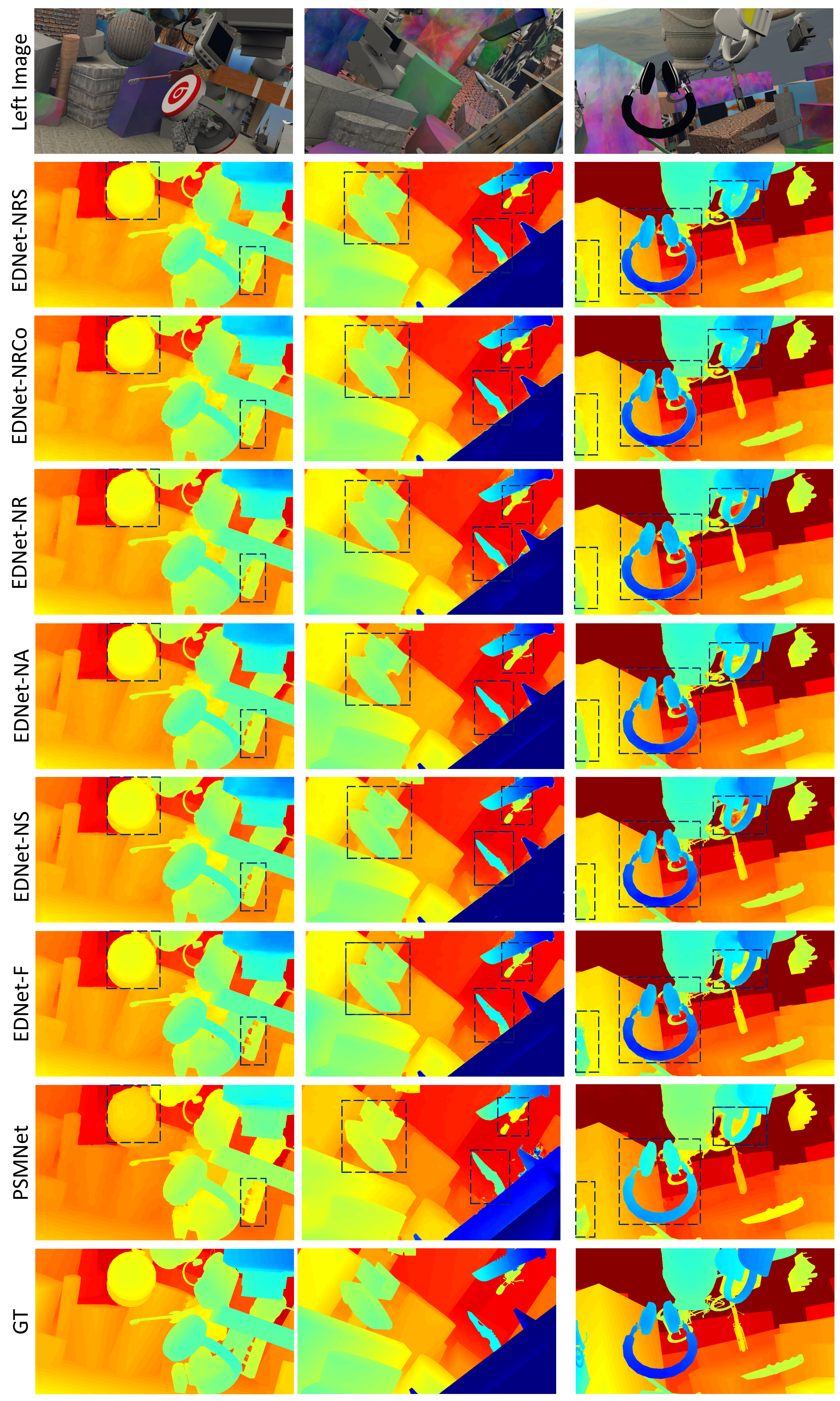}
	\caption{Visualization of the ablation study results on Scene Flow dataset. Our proposed combined volume and spatial residual module can bring great performance improvement especially at bounding box areas.}
	\label{fig:sf_ablation}
	\vspace{-0.8 em}
\end{figure}

\textbf{Cost Volume Combination:} As shown in Figure \ref{fig:sf_ablation}, models without the combined volume suffer from inaccurate and discontinuous disparity estimation, especially in low-texture regions. The possible reason might be that a single cost volume is unable to avoid the information loss which leads to less robust feature representations. As shown in Table \ref{tab:ablation}, the EPE decreases from EDNet-NS’s 1.07 to EDNet-F’s 1.00. The comparison among EDNet-NRS, EDNet-NRCo and EDNet-NR can validate the effectiveness of our proposed combined volume as well. 
\begin{figure}[htbp]
	\centering
	\includegraphics[width=0.96\linewidth]{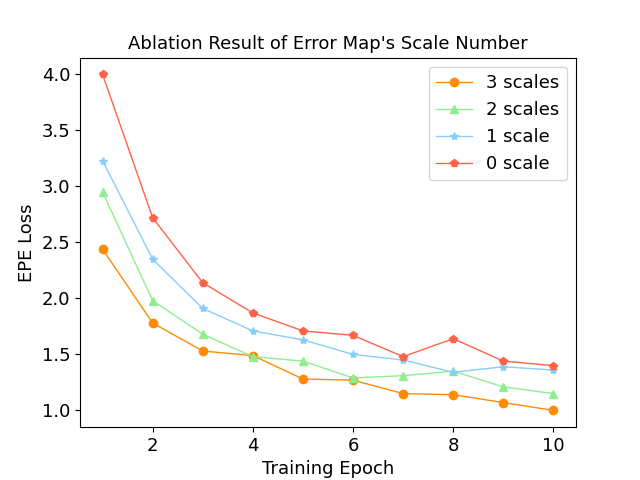}
	\caption{Comparison among models with different numbers of error maps. The multi-scale error maps speed up the convergence and achieve a lower loss. ``$n$ scale(s)'' indicates that  the error map is applied to the residual features from scale\_0 to scale\_$n$.}
	\label{fig:convergence}
	\vspace{-0.8 em}
\end{figure}

\textbf{Attention-based Spatial Residual:} It can be directly analyzed from EDNet-NR and EDNet-NA in Table \ref{tab:ablation} that the residual learning module brings significant improvement in terms of accuracy, about 36\% decrease of EPE. The comparison between EDNet-NA and EDNet-F in Table \ref{tab:ablation} demonstrates that the learning efficiency can be further improved with the help of the attention mechanism. The visualization of EDNet-NA and EDNet-F shown in Figure \ref{fig:sf_ablation} illustrates that more details like the shaper object boundaries can be recovered with the attention-based spatial residual.

\textbf{Multi-scale Error Maps:} Experiments are conducted to prove that our multi-scale error maps mechanism is of great importance to the residual learning. We remove the error map from the residual features as well as the input of attention module from scale 2 to all scale. As shown in Figure \ref{fig:convergence}, models with error maps at multiple scales can achieve a lower EPE loss with a faster convergence speed. Such performance gain comes at a low cost of extra computation. Multi-scale error maps enable the residual module to learn from the error at the corresponding scale and thus greatly explore the ability of residual learning.

\begin{table*}[htbp]
	\centering
	\begin{tabular}{ccccccccc} \hline
		Method & PSMNet \cite{psmnet} & GANet \cite{ganet} & GwcNet \cite{gwcnet} & Bi3D \cite{Bi3D} & DispNetC \cite{dispnetc} & AANet+ \cite{AANET} & Ours \\ \hline
		EPE & 1.09	& 0.84	& 0.76	& 0.73 & 1.68	& \underline{0.72} & \textbf{0.63} \\ \hline
		Time (s) & 0.453 & 3.302 & 0.254 & OOM & \textbf{0.025} & 0.068 & \underline{0.059} \\ \hline
	\end{tabular}
	\caption{EPE values on Scene Flow dataset of several state-of-the-art methods. Our method achieves the best scores and has a competitive inference speed. The inference time is tested on a single NVIDIA 2080ti GPU at the resolution of 576$\times$960 for a fair comparison. OOM denotes out of memory. ``\underline{X}'' indicates the second best.}
	\label{tab:epe_sota}
\end{table*}
\subsection{Experimental Results}
In this subsection, we compare our method with those existing state-of-the-art methods in the aspect of inference speed, memory consumption and accuracy on Scene Flow, KITTI 2015 and KITTI 2012 datasets. We also evaluate the model generalization on Middlebury 2014 dataset \cite{middlebury2014}.
\begin{figure}[htbp]
	\centering
	\includegraphics[width=0.75\linewidth]{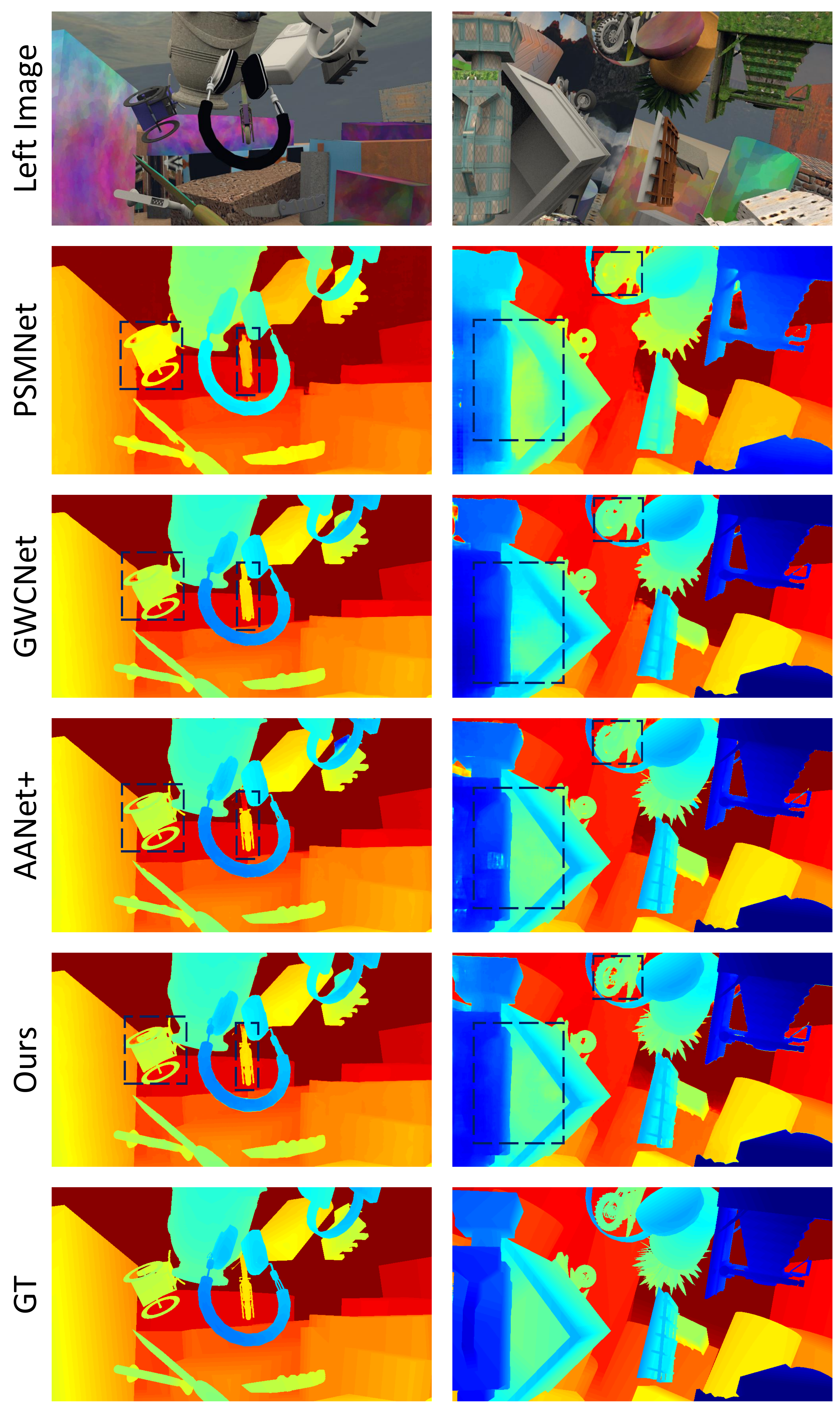}
	\caption{Visual comparison of our EDNet and some other top-performing methods. Please enlarge the bounding box areas for more details.}
	\label{fig:sf_evaluation}
	\vspace{-0.8 em}
\end{figure}
\begin{table}[htbp]
	\centering
	\begin{tabular}{cccccc} \hline
		\multirow{2}{*}{Method} & \multicolumn{2}{c}{Noc (\%)} & \multicolumn{2}{c}{All (\%)} & \multirow{2}{*}{Time (s)} \\ \cline{2-5}
		& fg & all & fg & all & \\ \hline
		GANet \cite{ganet} & 3.37 & \textbf{1.73} & 3.82 & 1.93 & 0.36 \\ 
		GCNet \cite{gcnet} & 5.58 & 2.61 & 6.16 & 2.87 & 0.9 \\ 
		PSMNet \cite{psmnet} & 4.31 & 2.14 & 4.62 & 2.32 & 0.41 \\ 
		GwcNet \cite{gwcnet} & 3.49 & 1.92 & 3.93 & 2.11 & 0.32 \\ 
		SegStereo \cite{segstereo} & 3.70 & 2.08 & 4.07 & 2.25 & 0.6 \\ 
		MC-CNN \cite{mccnn} & 7.64 & 3.33 & 8.88 & 3.89 & 67 \\
		HD$^3$ \cite{hd3} & 3.43 & 1.87 & 3.63 & 2.02 & 0.14 \\
		CSN \cite{CSN} & 3.55 & 1.78 & 4.03 & \textbf{1.59} & 0.6 \\ 
		DeepPruner-B \cite{deeppruner} & 3.18 & 1.95 & 3.56 & 2.15 & 0.18 \\
		Bi3D \cite{Bi3D} & \textbf{3.11} & 1.79 & \textbf{3.48} & 1.95 & 0.48 \\
		Ours  & 3.33 & 2.31 & 3.88 & 2.53 & \textbf{0.05} \\ \hline
		AANet \cite{AANET} & 4.93 & 2.32 & 5.39 & 2.55 & 0.075 \\
		DeepPruner-F \cite{deeppruner} & 3.43 & 2.35 & 3.91 & 2.59 & 0.06 \\
		DispNetC \cite{dispnetc} & 3.72 & 4.05 & 4.41 & 4.34 & 0.06 \\
		FADNet \cite{wang2020fadnet} & \textbf{3.07} & 2.59 & \textbf{3.50} & 2.82 & 0.05 \\
		MADNet \cite{madnet} & 8.41 & 4.27 & 9.20 & 4.66 & 0.02 \\
		Ours  & 3.33 & \textbf{2.31} & 3.88 & \textbf{2.53} & 0.05 \\ \hline
	\end{tabular}
    \caption{Benchmark results on KITTI 2015 test sets. “Noc” and “All” indicate the percentage of outliers averaged over ground truth pixels of non-occluded and all regions respectively. “fg” and “all” indicate the percentage of outliers averaged over the foreground and all ground truth pixels respectively. “\textbf{X}” indicates the best result.}
	\label{tab:kitti2015}
\end{table}
\begin{figure*}[htbp]
	\centering
	\includegraphics[width=0.96\linewidth]{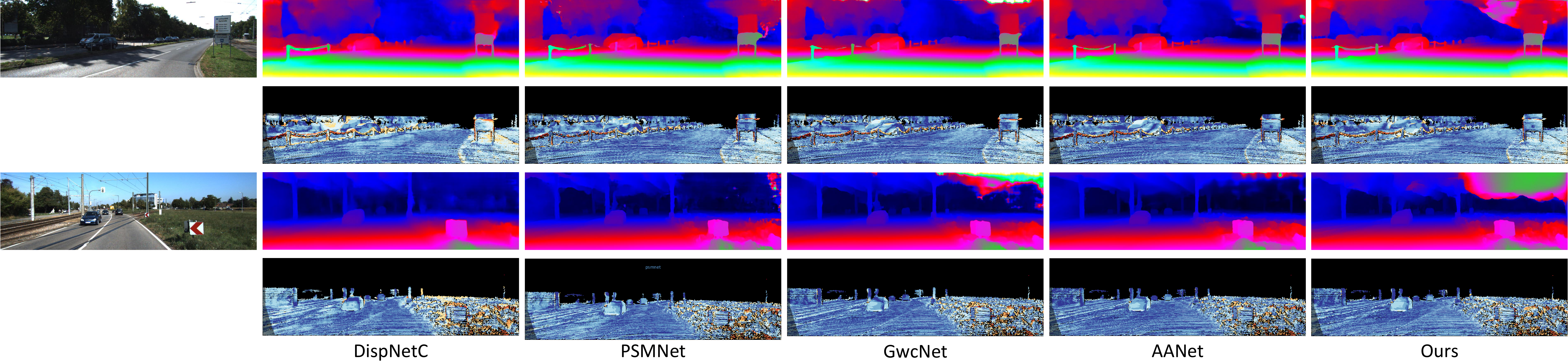}
	\caption{Results of the disparity prediction for KITTI 2015 testing data. The leftmost column shows left images of the stereo pairs. The rest five columns show the disparity maps predicted by DispNetC \cite{dispnetc}, PSMNet \cite{psmnet}, GwcNet \cite{gwcnet}, AANet\cite{AANET} and our EDNet, as well as their error maps.}
	\label{fig:kitti}
	\vspace{-0.8 em}
\end{figure*}

\textbf{Scene Flow Dataset:} As shown in Table \ref{tab:epe_sota},  our proposed EDNet not only outperforms all the competing state-of-the-art methods with the lowest EPE but also runs significantly faster: approximately 7.5$\times$ faster than PSMNet \cite{psmnet}, 4$\times$ faster than GwcNet \cite{gwcnet}, 55$\times$ faster than GANet \cite{ganet}. Compared with DispNetC \cite{dispnetc} and StereoNet \cite{stereonet}, EDNet improves the performance by 60\% and 40\% respectively. Figure \ref{fig:sf_evaluation} gives a visual comparison between our EDNet and other outstanding methods. Sharper object boundaries and more continuous disparity maps can be generated by EDNet, indicating the value of our proposed approach.

\textbf{KITTI Datasets:} We divide the comparison into two groups as shown in Table \ref{tab:kitti2015} and Table \ref{tab:kitti2012}. First, compared with other top-performing methods, our EDNet still achieves competitive results when evaluating non-occluded pixels while runs considerably faster according to the benchmark. Then we compare our EDNet with previous real-time models. Experimental results from Table \ref{tab:kitti2015} and Table \ref{tab:kitti2012} show that our work can produce more precise estimation.  To stress the efficiency of our proposed method, we compare the computational complexity, memory consumption as well as the inference speed with some popular 3D convolutions based models. Table \ref{tab:perf} comprehensively shows that our model requires less memory consumption: approximately 50\% less than PSMNet \cite{psmnet}, 40\% less than GwcNet \cite{gwcnet}, 60\% less than GANet \cite{ganet} and 70\% less than Bi3D \cite{Bi3D}. Moreover, our EDNet has lower computational complexity as our work has the lowest FLOPs (the lower the better) and runs significantly faster: approximately 7$\times$ faster than PSMNet \cite{psmnet}, 17$\times$ faster than Bi3D \cite{Bi3D} and 45$\times$ faster than GANet \cite{ganet}. Figure \ref{fig:kitti} visualizes the disparity and error maps of our method and other state-of-the-art works on KITTI 2015 dataset.
\begin{table}[htbp]
	\centering
	\begin{tabular}{cccccc} \hline
		\multirow{2}{*}{Method} & \multicolumn{2}{c}{Out (\%)} & \multicolumn{2}{c}{Avg} & \multirow{2}{*}{Time (s)} \\ \cline{2-5}
		& Noc & All & Noc & All & \\ \hline
		GANet \cite{ganet} & 2.18 & 2.79 & \textbf{0.5} & \textbf{0.5} & 0.36 \\ 
		GCNet \cite{gcnet} & 2.71 & 3.46 & 0.6 & 0.7 & 0.90 \\ 
		PSMNet \cite{psmnet} & 2.44 & 3.01 & \textbf{0.5} & 0.6 & 0.41 \\ 
		GwcNet \cite{gwcnet} & \textbf{2.16} & \textbf{2.71} & \textbf{0.5} & \textbf{0.5} & 0.32 \\ 
		SegStereo \cite{segstereo} & 2.66 & 3.19 & \textbf{0.5} & 0.6 & 0.60 \\ 
		MC-CNN \cite{mccnn} & 3.90 & 5.45 & 0.8 & 1.0 & 100 \\
		Ours  & 2.97 & 3.67 & \textbf{0.5} & 0.6 & \textbf{0.05} \\ \hline
		AANet \cite{AANET} & \textbf{2.90} & \textbf{3.60} & \textbf{0.5} & \textbf{0.6} & 0.06 \\
		StereoNet \cite{stereonet} & 4.91 & 6.02 & 0.8 & 0.9 & \textbf{0.015} \\
		DispNetC \cite{dispnetc} & 7.36 & 8.70 & 0.9 & 1.0 & 0.06 \\
		FADNet \cite{wang2020fadnet} & 3.98 & 4.61 & 0.6 & 0.7 & 0.05 \\
		Ours  & 2.97 & 3.67 & \textbf{0.5} & \textbf{0.6} & 0.05 \\ \hline
	\end{tabular}
	\caption{Benchmark results on KITTI2012 test sets. Both the percentage of pixels with error bigger than 2 and the overall EPE are reported over non-occluded and all regions.}
	\label{tab:kitti2012}
\end{table}
\begin{table}[htbp]
	\centering
	\begin{tabular}{cccc} \hline
		Method & Memory (GB) & GFLOPs & Time (s) \\ \hline
		PSMNet \cite{psmnet} & 4.83 & 937.9 & 0.393 \\ \hline
		GANet \cite{ganet} & 6.53 & 1936.98 & 2.43 \\ \hline
		GwcNet \cite{gwcnet} & 4.27 & 899.99 & 0.272 \\ \hline
		Bi3D \cite{Bi3D} & 10.74 & 4212.05 & 0.899 \\ \hline
		Ours & \textbf{2.52} & \textbf{162.92} & \textbf{0.053} \\ \hline
	\end{tabular}
	\caption{Comparisons of the runtime, running memory and computational cost. All the results are tested on a single NVIDIA RTX 2080Ti GPU 
	at a resolution of 1248$\times$384.}
	\label{tab:perf}
	\vspace{-1.0 em}
\end{table}
\begin{figure}[htbp]
	\centering
	\includegraphics[width=0.96\linewidth]{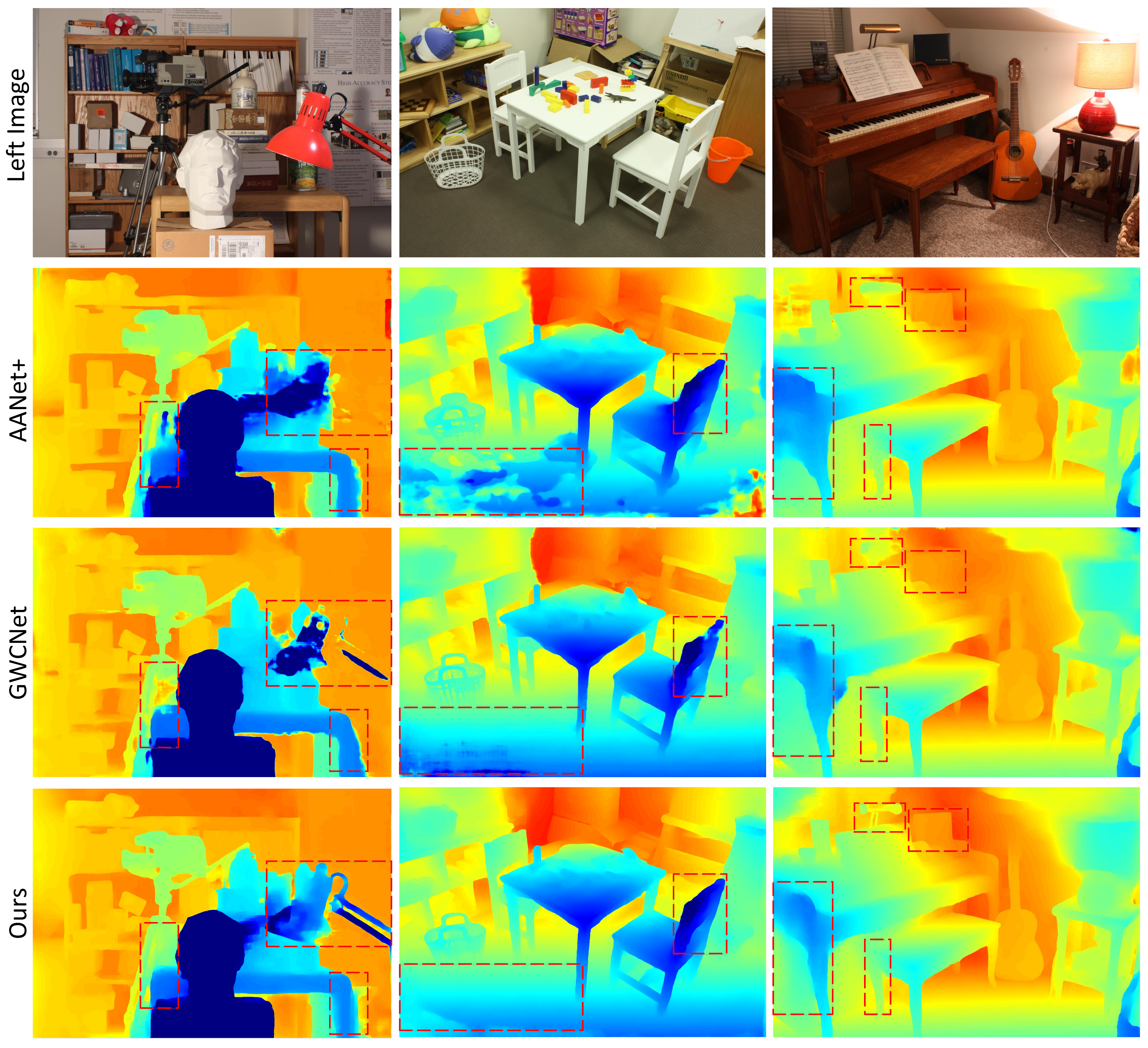}
	\caption{Evaluation of model generalization on Middlebury 2014 dataset. Our EDNet produces more smooth and continous disparity maps. Sharper object boundaries as well as better overall structures can be recovered. Please zoom in bounding box areas for a further comparison.}
	\label{fig:mb}
\end{figure}

\textbf{Middlebury 2014 Dataset:} We evaluate the model generalization on Middlebury 2014 dataset without additional training. All the models are pretrained on Scene Flow and finetuned on KITTI. As observed from Figure \ref{fig:mb}, the performance of our EDNet is superior to others, as more smooth and continuous disparity estimation in low-texture regions can be produced. Moreover, our EDNet can preserve better overall object structures and generate sharper edges. The generalization results show that our model has a better adaptation ability to unknown scenes.

\section{Conclusion}\label{sec:conclusion}
We have exploited an efficient architecture called EDNet with the proposed combined volume and attention-based residual module. We show that the combination of the correlation volume and squeezed 4D concatenation volume is of great importance to robust feature representations, especially in ill-posed regions. Besides, the employment of the spatial attention module greatly improves the efficiency of residual learning with multi-scale error maps. Extensive experimental results on  KITTI and Scene Flow datasets have demonstrated the superiority of our method when comparing with previous state-of-the-art methods. The future work would be applying our work to other depth-related tasks, e.g., 3D reconstruction and robot navigation.

{\small
\bibliographystyle{IEEEtran}
\bibliography{main}

\begin{thebibliography}{10}
\providecommand{\url}[1]{#1}
\csname url@samestyle\endcsname
\providecommand{\newblock}{\relax}
\providecommand{\bibinfo}[2]{#2}
\providecommand{\BIBentrySTDinterwordspacing}{\spaceskip=0pt\relax}
\providecommand{\BIBentryALTinterwordstretchfactor}{4}
\providecommand{\BIBentryALTinterwordspacing}{\spaceskip=\fontdimen2\font plus
\BIBentryALTinterwordstretchfactor\fontdimen3\font minus
  \fontdimen4\font\relax}
\providecommand{\BIBforeignlanguage}[2]{{%
\expandafter\ifx\csname l@#1\endcsname\relax
\typeout{** WARNING: IEEEtran.bst: No hyphenation pattern has been}%
\typeout{** loaded for the language `#1'. Using the pattern for}%
\typeout{** the default language instead.}%
\else
\language=\csname l@#1\endcsname
\fi
#2}}
\providecommand{\BIBdecl}{\relax}
\BIBdecl

\bibitem{gcnet}
A.~{Kendall}, H.~{Martirosyan}, S.~{Dasgupta}, P.~{Henry}, R.~{Kennedy},
  A.~{Bachrach}, and A.~{Bry}, ``End-to-end learning of geometry and context
  for deep stereo regression,'' in \emph{2017 IEEE International Conference on
  Computer Vision (ICCV)}, 2017, pp. 66--75.

\bibitem{psmnet}
J.~{Chang} and Y.~{Chen}, ``Pyramid stereo matching network,'' in \emph{2018
  IEEE/CVF Conference on Computer Vision and Pattern Recognition}, 2018, pp.
  5410--5418.

\bibitem{gwcnet}
X.~{Guo}, K.~{Yang}, W.~{Yang}, X.~{Wang}, and H.~{Li}, ``Group-wise
  correlation stereo network,'' in \emph{2019 IEEE/CVF Conference on Computer
  Vision and Pattern Recognition (CVPR)}, 2019, pp. 3268--3277.

\bibitem{ganet}
F.~{Zhang}, V.~{Prisacariu}, R.~{Yang}, and P.~H.~S. {Torr}, ``Ga-net: Guided
  aggregation net for end-to-end stereo matching,'' in \emph{2019 IEEE/CVF
  Conference on Computer Vision and Pattern Recognition (CVPR)}, 2019.

\bibitem{dispnetc}
N.~{Mayer}, E.~{Ilg}, P.~{Häusser}, P.~{Fischer}, D.~{Cremers},
  A.~{Dosovitskiy}, and T.~{Brox}, ``A large dataset to train convolutional
  networks for disparity, optical flow, and scene flow estimation,'' in
  \emph{2016 IEEE Conference on Computer Vision and Pattern Recognition
  (CVPR)}, 2016, pp. 4040--4048.

\bibitem{consistency}
Z.~{Liang}, Y.~{Feng}, Y.~{Guo}, H.~{Liu}, W.~{Chen}, L.~{Qiao}, L.~{Zhou}, and
  J.~{Zhang}, ``Learning for disparity estimation through feature constancy,''
  in \emph{2018 IEEE/CVF Conference on Computer Vision and Pattern
  Recognition}, 2018, pp. 2811--2820.

\bibitem{segstereo}
G.~Yang, H.~Zhao, J.~Shi, Z.~Deng, and J.~Jia, ``Segstereo: Exploiting semantic
  information for disparity estimation,'' in \emph{Proceedings of the European
  Conference on Computer Vision (ECCV)}, September 2018.

\bibitem{CRL}
J.~{Pang}, W.~{Sun}, J.~S. {Ren}, C.~{Yang}, and Q.~{Yan}, ``Cascade residual
  learning: A two-stage convolutional neural network for stereo matching,'' in
  \emph{2017 IEEE International Conference on Computer Vision Workshops
  (ICCVW)}, 2017, pp. 878--886.

\bibitem{resnet}
K.~{He}, X.~{Zhang}, S.~{Ren}, and J.~{Sun}, ``Deep residual learning for image
  recognition,'' in \emph{2016 IEEE Conference on Computer Vision and Pattern
  Recognition (CVPR)}, 2016, pp. 770--778.

\bibitem{wang2020fadnet}
Q.~Wang, S.~Shi, S.~Zheng, K.~Zhao, and X.~Chu, ``{FADNet}: A fast and accurate
  network for disparity estimation,'' in \emph{2020 {IEEE} International
  Conference on Robotics and Automation ({ICRA} 2020)}, 2020, pp. 101--107.

\bibitem{edgestereo}
X.~{Song}, X.~{Zhao}, H.~{Hu}, and L.~{Fang}, ``Edgestereo: A context
  integrated residual pyramid network for stereo matching,'' in \emph{2019
  Asian Conference on Computer Vision (ACCV)}, 2018.

\bibitem{AANET}
H.~{Xu} and J.~{Zhang}, ``Aanet: Adaptive aggregation network for efficient
  stereo matching,'' in \emph{2020 IEEE/CVF Conference on Computer Vision and
  Pattern Recognition (CVPR)}, 2020, pp. 1956--1965.

\bibitem{madnet}
A.~{Tonioni}, F.~{Tosi}, M.~{Poggi}, S.~{Mattoccia}, and L.~D. {Stefano},
  ``Real-time self-adaptive deep stereo,'' in \emph{2019 IEEE/CVF Conference on
  Computer Vision and Pattern Recognition (CVPR)}, 2019, pp. 195--204.

\bibitem{geonet}
Z.~{Yin} and J.~{Shi}, ``Geonet: Unsupervised learning of dense depth, optical
  flow and camera pose,'' in \emph{2018 IEEE/CVF Conference on Computer Vision
  and Pattern Recognition}, 2018, pp. 1983--1992.

\bibitem{flownet2}
E.~{Ilg}, N.~{Mayer}, T.~{Saikia}, M.~{Keuper}, A.~{Dosovitskiy}, and
  T.~{Brox}, ``Flownet 2.0: Evolution of optical flow estimation with deep
  networks,'' in \emph{2017 IEEE Conference on Computer Vision and Pattern
  Recognition (CVPR)}, 2017, pp. 1647--1655.

\bibitem{KITTI2015}
M.~{Menze} and A.~{Geiger}, ``Object scene flow for autonomous vehicles,'' in
  \emph{2015 IEEE Conference on Computer Vision and Pattern Recognition
  (CVPR)}, 2015, pp. 3061--3070.

\bibitem{KITTI2012}
A.~{Geiger}, P.~{Lenz}, and R.~{Urtasun}, ``Are we ready for autonomous
  driving? the kitti vision benchmark suite,'' in \emph{2012 IEEE Conference on
  Computer Vision and Pattern Recognition}, 2012, pp. 3354--3361.

\bibitem{taxonomy2001}
D.~{Scharstein}, R.~{Szeliski}, and R.~{Zabih}, ``A taxonomy and evaluation of
  dense two-frame stereo correspondence algorithms,'' in \emph{Proceedings IEEE
  Workshop on Stereo and Multi-Baseline Vision (SMBV 2001)}, 2001.

\bibitem{LeCun2016}
J.~{\v{Z}}bontar and Y.~LeCun, ``Stereo matching by training a convolutional
  neural network to compare image patches,'' \emph{Journal of Machine Learning
  Research}, vol.~17, no.~65, pp. 1--32, 2016.

\bibitem{flownet}
A.~{Dosovitskiy}, P.~{Fischer}, E.~{Ilg}, P.~{Häusser}, C.~{Hazirbas},
  V.~{Golkov}, P.~v.~d. {Smagt}, D.~{Cremers}, and T.~{Brox}, ``Flownet:
  Learning optical flow with convolutional networks,'' in \emph{2015 IEEE
  International Conference on Computer Vision (ICCV)}, 2015, pp. 2758--2766.

\bibitem{embedding2015}
Z.~{Chen}, X.~{Sun}, L.~{Wang}, Y.~{Yu}, and C.~{Huang}, ``A deep visual
  correspondence embedding model for stereo matching costs,'' in \emph{2015
  IEEE International Conference on Computer Vision (ICCV)}, 2015, pp. 972--980.

\bibitem{feng2017}
Y.~{Feng}, Z.~{Liang}, and H.~{Liu}, ``Efficient deep learning for stereo
  matching with larger image patches,'' in \emph{2017 10th International
  Congress on Image and Signal Processing, BioMedical Engineering and
  Informatics (CISP-BMEI)}, 2017.

\bibitem{SPP}
K.~{He}, X.~{Zhang}, S.~{Ren}, and J.~{Sun}, ``Spatial pyramid pooling in deep
  convolutional networks for visual recognition,'' \emph{IEEE Transactions on
  Pattern Analysis and Machine Intelligence}, vol.~37, no.~9, pp. 1904--1916,
  2015.

\bibitem{MCUS}
G.~{Nie}, M.~{Cheng}, Y.~{Liu}, Z.~{Liang}, D.~{Fan}, Y.~{Liu}, and Y.~{Wang},
  ``Multi-level context ultra-aggregation for stereo matching,'' in \emph{2019
  IEEE/CVF Conference on Computer Vision and Pattern Recognition (CVPR)}, 2019,
  pp. 3278--3286.

\bibitem{densenet}
G.~{Huang}, Z.~{Liu}, L.~{Van Der Maaten}, and K.~Q. {Weinberger}, ``Densely
  connected convolutional networks,'' in \emph{2017 IEEE Conference on Computer
  Vision and Pattern Recognition (CVPR)}, 2017, pp. 2261--2269.

\bibitem{stereonet}
S.~Khamis, S.~Fanello, C.~Rhemann, A.~Kowdle, J.~Valentin, and S.~Izadi,
  ``Stereonet: Guided hierarchical refinement for real-time edge-aware depth
  prediction,'' in \emph{Proceedings of the European Conference on Computer
  Vision (ECCV)}, September 2018.

\bibitem{resdepth}
C.~{Stucker} and K.~{Schindler}, ``Resdepth: Learned residual stereo
  reconstruction,'' in \emph{2020 IEEE/CVF Conference on Computer Vision and
  Pattern Recognition Workshops (CVPRW)}, 2020, pp. 707--716.

\bibitem{unet}
O.~{Ronneberger}, P.~{Fischer}, and T.~{Brox}, ``U-net:convolutional networks
  for biomedical image segmentation,'' in \emph{2015 Medical Image Computing
  and Computer-Assisted Intervention (MICCAI)}, 2015.

\bibitem{anytime}
Y.~{Wang}, Z.~{Lai}, G.~{Huang}, B.~H. {Wang}, L.~{van der Maaten},
  M.~{Campbell}, and K.~Q. {Weinberger}, ``Anytime stereo image depth
  estimation on mobile devices,'' in \emph{2019 International Conference on
  Robotics and Automation (ICRA)}, 2019, pp. 5893--5900.

\bibitem{pytorch}
A.~Paszke, S.~Gross, F.~Massa, A.~Lerer, J.~Bradbury, G.~Chanan, T.~Killeen,
  Z.~Lin, N.~Gimelshein, L.~Antiga, A.~Desmaison, A.~Kopf, E.~Yang, Z.~DeVito,
  M.~Raison, A.~Tejani, S.~Chilamkurthy, B.~Steiner, L.~Fang, J.~Bai, and
  S.~Chintala, ``Pytorch: An imperative style, high-performance deep learning
  library,'' in \emph{Advances in Neural Information Processing Systems}, 2019,
  pp. 8026--8037.

\bibitem{nature2003}
M.~D. {Menz} and R.~D. {Freeman}, ``Stereoscopic depth processing in the visual
  cortex: a coarse-to-fine mechanism.'' in \emph{2003 Nature neuroscience},
  2003, pp. 59--65.

\bibitem{ImageNet}
J.~{Deng}, W.~{Dong}, R.~{Socher}, L.~{Li}, {Kai Li}, and {Li Fei-Fei},
  ``Imagenet: A large-scale hierarchical image database,'' in \emph{2009 IEEE
  Conference on Computer Vision and Pattern Recognition}, 2009, pp. 248--255.

\bibitem{Bi3D}
A.~{Badki}, A.~{Troccoli}, K.~{Kim}, J.~{Kautz}, P.~{Sen}, and O.~{Gallo},
  ``Bi3d: Stereo depth estimation via binary classifications,'' in \emph{2020
  IEEE/CVF Conference on Computer Vision and Pattern Recognition (CVPR)}, 2020,
  pp. 1597--1605.

\bibitem{middlebury2014}
D.~{Scharstein}, H.~{Hirschmuller}, Y.~{Kitajima}, G.~{Krathwohl}, N.~{Nesic},
  and P.~W. X~{Wang}, ``High-resolution stereo datasets with subpixel-accurate
  ground truth.'' in \emph{German Conference on Pattern Recognition (GCPR)},
  2014.

\bibitem{mccnn}
J.~{Žbontar} and Y.~{LeCun}, ``Computing the stereo matching cost with a
  convolutional neural network,'' in \emph{2015 IEEE Conference on Computer
  Vision and Pattern Recognition (CVPR)}, 2015, pp. 1592--1599.

\bibitem{hd3}
Z.~{Yin}, T.~{Darrell}, and F.~{Yu}, ``Hierarchical discrete distribution
  decomposition for match density estimation,'' in \emph{2019 IEEE/CVF
  Conference on Computer Vision and Pattern Recognition (CVPR)}, 2019, pp.
  6037--6046.

\bibitem{CSN}
X.~{Gu}, Z.~{Fan}, S.~{Zhu}, Z.~{Dai}, F.~{Tan}, and P.~{Tan}, ``Cascade cost
  volume for high-resolution multi-view stereo and stereo matching,'' in
  \emph{2020 IEEE/CVF Conference on Computer Vision and Pattern Recognition
  (CVPR)}, 2020, pp. 2492--2501.

\bibitem{deeppruner}
S.~{Duggal}, S.~{Wang}, W.~{Ma}, R.~{Hu}, and R.~{Urtasun}, ``Deeppruner:
  Learning efficient stereo matching via differentiable patchmatch,'' in
  \emph{2019 IEEE/CVF International Conference on Computer Vision (ICCV)},
  2019, pp. 4383--4392.

\end{thebibliography}
}

\end{document}